# Nhận dạng 26 bậc tự do của bàn tay sử dụng phương pháp mô hình với ảnh màu – độ sâu

# Recognition of 26 Degrees of Freedom of Hands Using Model-based approach and Depth-Color Images

Quách Công Hoàng, Phạm Minh Triển, Đặng Anh Việt, Phạm Đình Tuân, Trần Thuận Hoàng, Phùng Mạnh Dương

Trường Đại học Công nghệ, Đại học Quốc gia Hà Nội

Hà Nội, Việt Nam

Email: duongpm@vnu.edu.vn

*Tóm tắt*— Trong báo cáo này, chúng tôi trình bày hướng tiếp cận mô hình để nhận dạng đầy đủ 26 bậc tự do của bàn tay. Thông tin vào bao gồm ảnh quan sát màu – độ sâu (RGB-D) thu được từ cảm biến ảnh Kinect và ảnh mô hình 3 chiều của bàn tay xây dựng từ cấu trúc giải phẫu học và ma trận đồ họa. Một hàm mục tiêu sau đó được thiết lập sao cho hàm đạt giá trị cực tiểu khi ảnh mô hình và ảnh quan sát là trùng nhau. Để giải bài toán tối ưu 26 chiều này, giải thuật tối ưu bầy đàn (PSO) được sử dụng với một số cải tiến. Đồng thời, những tác vụ đòi hỏi tính toán lớn được chúng tôi đưa vào khối xử lý đồ họa GPU để thực hiện tính toán song song. Kết quả mô phỏng và thực nghiệm cho thấy hệ thống có thể nhận dạng được 26 bậc tự do của bàn tay với tốc độ xử lý 0,8 giây/hình. Giải thuật kém nhạy với nhiễu môi trường. Phần cứng yêu cầu đơn giản với một camera.

Abstract— In this study, we present an model-based approach to recognize full 26 degrees of freedom of a human hand. Input data include RGB-D images acquired from a Kinect camera and a 3D model of the hand constructed from its anatomy and graphical matrices. A cost function is then defined so that its minimum value is achieved when the model and observation images are matched. To solve the optimization problem in 26 dimensional space, the particle swarm optimization algorimth with improvements are used. In addition, parallel computation in graphical processing units (GPU) is utilized to handle computationally expensive tasks. Simulation and experimental results show that the system can recognize 26 degrees of freedom of hands with the processing time of 0.8 seconds per frame. The algorithm is robust to noise and the hardware requirement is simple with a single camera.

*Từ khóa*—nhận dạng bàn tay, giải thuật bầy đàn, cảm biến ảnh Kinect

## I. GIỚI THIỆU

Khi máy tính ngày càng thu nhỏ kích thước như một chiếc kính hay chiếc đồng hồ đeo tay thì việc sử dụng bàn phím, chuột hay màn hình cảm ứng trở nên không thích hợp. Thay vào đó, những cách thức tương tác người – máy mới cần được thúc đẩy nghiên cứu [17]. Bàn tay, bộ phận hoạt động chính xác và hiệu quả nhất khi con người sử dụng công cụ, được đánh giá nhiều tiềm năng. Và thực tế bài toán nhận dạng tư thế tay đã nhận được nhiều sự quan tâm nghiên cứu và đã có những ứng dụng cụ thể như tương tác robot, nhận diện ngôn ngữ cử chỉ, hay điều khiển thiết bị [1]… Tuy nhiên, các ứng dụng tương tác hiện đại như thực tế ảo (VR) và thực tế tăng cường (AR) thường đòi hỏi độ chính xác cao cùng số bậc tự do lớn khiến các phương pháp truyền thống tỏ ra kém hiệu quả. Thay vào đó, phương pháp mô hình được xem là hướng tiếp cận khả thi hiện nay [1] – [4].

Phương pháp mô hình nhận dạng dựa trên so sánh *ảnh quan sát* với *ảnh mô hình* 3 chiều. Ảnh quan sát là hình ảnh thu được từ hệ một hoặc nhiều camera và có thể kèm thông tin độ sâu. Trong khi đó, ảnh mô hình được xây dựng dựa trên cấu trúc giải phẫu học cùng các ma trận đồ họa. Tùy mục đích ứng dụng và giải thuật, ảnh mô hình có thể khác nhau giữa các nhóm nghiên cứu.

Trong [3], mô hình bàn tay được xây gồm 12 bậc tự do với 10 bậc dành cho các ngón tay và 2 bậc dành cho chuyển động tịnh tiến trong không gian. Để nhận dạng tư thế tay, có hai phép đo được sử dụng. Phép đo thứ nhất đo mức độ chồng chập về diện tích giữa ảnh quan sát và ảnh mô hình chiếu trên mặt phẳng quan sát. Phép đo thứ hai đánh giá sự sai khác về khoảng cách giữa các đường biên của hai ảnh. Kĩ thuật tối ưu xuống dốc đơn hình (downhill simplex) sau đó được sử dụng để tìm tư thế cho sai khác nhỏ nhất. Các ràng buộc cơ sinh học cũng được sử dụng nhằm thu hẹp không gian tìm kiếm và loại bỏ các trường hợp không thực. Kết quả thực nghiệm cho thấy giải thuật đã nhận dạng được chuyển động đơn giản của bàn tay trong điều kiện nền đồng màu.

Trong một nghiên cứu khác [4], Stenger đề xuất mô hình bàn tay với 27 bậc tự do được biểu diễn bởi 39 mặt bậc hai cụt. Việc sử dụng mặt bậc hai giúp đơn giản quá trình khởi tạo mô hình 3 chiều đồng thời dễ dàng thực hiện các phép chiếu. Bộ lọc Kalman sau đó được sử dụng để ước lượng và tối thiểu sai số hình học giữa các đường biên của ảnh quan sát và ảnh mô hình. Kết quả cho thấy giải thuật có thể nhận dạng được 7 bậc tự do với tốc độ 3 hình/giây. Để nâng cao độ chính xác,

Stenger sau đó đã đề xuất sử dụng tập hợp mẫu gồm 16.055 tư thế bàn tay kết hợp với bộ lọc Bayes phân cấp [5]. Các hàm so sánh tương quan cũng được cải tiến để có thể làm việc được điều kiện nhiễu môi trường lớn. Giải thuật thành công với tỉ lệ nhận dạng hơn 90% và độ chính xác 9.3 điểm ảnh cho ảnh 320x240. Tuy nhiên, quá trình cài đặt thuật toán tương đối phức tạp với nhiều bước căn chỉnh thủ công đồng thời yêu cầu phải có tập dữ liệu quan sát lớn.

Gần đây, Oikonomidis đã đề xuất mô hình bàn tay gồm 26 bậc tự do được xây dựng từ các hình đồ họa cơ bản là hình cầu, hình trụ và hình elipsoid [2]. Ảnh quan sát được sử dụng bao gồm ảnh màu và ảnh độ sâu thu thập bởi cảm biến ảnh Kinect. Giải thuật tối ưu bầy đàn sau đó được áp dụng để tìm nghiệm cho bài toán cực tiểu sự sai khác giữa ảnh quan sát và ảnh mô hình. Kết quả cho thấy giải thuật đã nhận diện được đầy đủ 26 bậc tự do của bàn tay với tốc độ 15 hình/giây. Tuy vậy, quá trình khởi tạo ban đầu vẫn phải thực hiện thủ công.

Trong báo cáo này, chúng tôi tiếp cận theo hướng mô hình để giải quyết bài toán nhận dạng tư thế bàn tay, hay cụ thể hơn là trạng thái các khớp nối của bàn tay. Vấn đề nhận dạng được xây dựng như một bài toán tối ưu với mục tiêu là tối thiểu sự sai khác giữa ảnh mô hình của các thế tư thế tay giả định với ảnh quan sát thu được từ cảm biến ảnh Kinect. Giải thuật bầy đàn cải tiến sau đó được sử dụng để giải bài toán tối ưu này. Đồng thời, các tác vụ đòi hỏi tính toán lớn được đưa vào khối xử lý đồ họa GPU của máy tính để tính toán song song. Kết quả thực nghiệm hiện tại cho thấy hệ thống có thể nhận dạng được 26 bậc tự do của bàn tay trong thời gian 0.8s. Kết quả nhận dạng kém nhạy với nhiễu môi trường và yêu cầu phần cứng đơn giản.

## II. MÔ HÌNH BÀN TAY

Để triển khai giải thuật nhận dạng, mô hình bàn tay bao gồm ảnh mô hình và ảnh quan sát cần được định nghĩa.

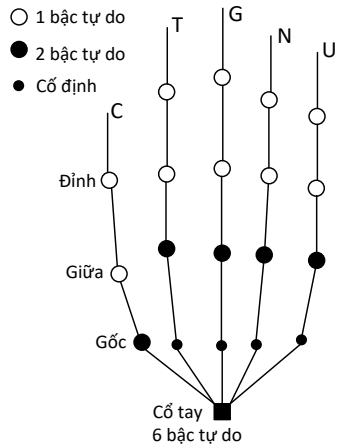

Hình 1. Mô hình động học của bàn tay

### A. Ảnh mô hình bàn tay

Bàn tay con người bao gồm 27 xương, trong đó có 8 xương ở cổ tay và 19 xương cho lòng bàn tay và ngón tay. Các xương này được kết nối với nhau bởi các khớp nối có một hoặc nhiều bậc tự do. Hình 1 biểu diễn các khớp nối cùng số bậc tự do tương ứng tạo thành tổng cộng 26 bậc tự do [1]. Trong đó, cổ tay có 6 bậc tự do với 3 bậc tự do cho chuyển động tịnh tiến trong không gian và 3 bậc tự do cho chuyển động xoay quanh các trục. Năm ngón tay mỗi ngón có 4 bậc tự do với 2 bậc cho khớp gốc ngón tay (gập/ngửa và khép/mở) và 1 bậc cho mỗi khớp còn lại.

Với cách biểu diễn như vậy, động học của mỗi ngón tay được xác định bởi một vector gồm 4 tham số góc:

$$q_i = (\theta_{MP}^x, \theta_{MP}^z, \theta_{PIP}, \theta_{DIP}) \quad (1)$$

trong đó $\theta_{MP}^x$ và $\theta_{MP}^z$ là hai góc quay của khớp gốc, $\theta_{PIP}$ là góc quay của khớp giữa và $\theta_{DIP}$ là góc quay của khớp đỉnh.

Tương tự, vị trí và hướng của bàn tay được xác định qua cổ tay bởi vector gồm 6 tham số:

$$q_c = (x_c, y_c, z_c, \theta_c^x, \theta_c^y, \theta_c^z) \quad (2)$$

trong đó $(x_c, y_c, z_c)$ là tọa độ của cổ tay trong không gian và $(\theta_c^x, \theta_c^y, \theta_c^z)$ là hướng của bàn tay quay quanh các trục tương ứng. Như vậy, tư thế của bàn tay hoàn toàn xác định khi biết 26 tham số góc:

$$h = (q_i, q_c), i = 1, 2, ..., 5 \quad (3)$$

Do đặc điểm giải phẫu học, chuyển động của các khớp ngón tay bị ràng buộc bởi các cơ giằng dẫn tới các góc quay của cổ tay và các đốt ngón tay bị giới hạn. Đặc điểm này là quan trọng bởi nó giúp giới hạn đáng kể không gian tìm kiếm của giải thuật bầy đàn sau này. Bảng 1 trình bày giới hạn của các tham số góc của ngón tay. Bảng 2 trình bày giới hạn các tham số của cổ tay. Lưu ý rằng giới hạn của vị trí $(x_c, y_c, z_c)$ được xác định bởi thị trường của camera.

BẢNG 1: GIỚI HẠN CÁC THAM SỐ GÓC CỦA NGÓN TAY

|  | $\theta_{MP}^x$ | $\theta_{MP}^z$ | $\theta_{PIP}$ | $\theta_{DIP}$ |
|---|---|---|---|---|
| Ngón cái | $0^0 - 90^0$ | $-15^0 - 60^0$ | $0^0 - 50^0$ | $-15^0 - 70^0$ |
| Ngón trỏ | $0^0 - 90^0$ | $-15^0 - 15^0$ | $0^0 - 100^0$ | $0^0 - 60^0$ |
| Ngón giữa | $0^0 - 90^0$ | $-10^0 - 10^0$ | $0^0 - 100^0$ | $0^0 - 60^0$ |
| Ngón đeo nhẫn | $0^0 - 90^0$ | $-30^0 - 0^0$ | $0^0 - 100^0$ | $0^0 - 60^0$ |
| Ngón út | $0^0 - 90^0$ | $-45^0 - 0^0$ | $0^0 - 100^0$ | $0^0 - 60^0$ |

BẢNG 2: GIỚI HẠN CÁC THAM SỐ GÓC VÀ VỊ TRÍ CỦA CỔ TAY

| $x_c$ | $y_c$ | $z_c$ | $\theta_c^x$ | $\theta_c^y$ | $\theta_c^z$ |
|---|---|---|---|---|---|
| -0,9 m – 0,9 m | -0,68 m – 0,68 m | 0,5 m – 1,5 m | $-30^0 - 120^0$ | $-70^0 - 75^0$ | $-35^0 - 20^0$ |

Từ cấu trúc giải phẫu học và động học, chúng tôi biểu diễn ảnh mô hình của bàn tay gồm 2 phần: lòng bàn tay và năm ngón tay. Lòng bàn tay được biểu diễn bởi một hình trụ elip bao hai đầu là 2 khối ellipsoid (hình 2). Mỗi ngón tay được biểu diễn bởi 3 hình nón cụt tương ứng với các đốt ngón tay và 4 hình cầu tương ứng với các khớp ngón tay và đầu ngón tay. Riêng ngón cái có cấu tạo hơi khác nên đốt ngón tay lớn nhất được biểu diễn bởi một khối ellipsoid thay vì hình nón cụt. Kích thước và tỉ lệ giữa các phần của bàn tay được xác định dựa trên đo đạc bàn tay thực. Trên máy tính, ảnh mô hình được chúng tôi biểu diễn đồ họa dựa trên thư viện OpenGL [12].

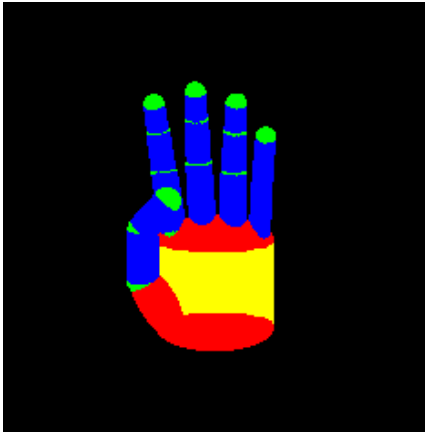

Hình 2. Ảnh mô hình bàn tay tạo bởi các khối hình học cơ bản

Ảnh mô hình cho phép biểu diễn hình ảnh 3 chiều của bàn tay trong không gian. Bằng các phép chiếu hình học lên mặt phẳng quan sát, ta có thể thu được ảnh màu và ảnh độ sâu. Các ảnh này được dùng để so sánh với ảnh quan sát thu từ cảm biến ảnh Kinect.

### B. Ảnh quan sát bàn tay

Ảnh quan sát là ảnh thu được từ một hoặc nhiều camera ghi lại hình ảnh bàn tay. Trong hệ thống của chúng tôi, ảnh quan sát bao gồm ảnh màu RGB và ảnh độ sâu ghi bởi một cảm biến ảnh Kinect [16] có độ phân giải 640x480 và tốc độ 15 hình/giây (hình 3). Bằng thuật toán nhận diện màu da và phân hoạch độ sâu, vùng bàn tay trên ảnh màu và ảnh độ sâu được trích chọn. Kết quả của giai đoạn tiền xử lý này sẽ cho ta ảnh quan sát $O = (O_s, O_d)$ với $O_s$ là ảnh màu và $O_d$ là ảnh độ sâu.

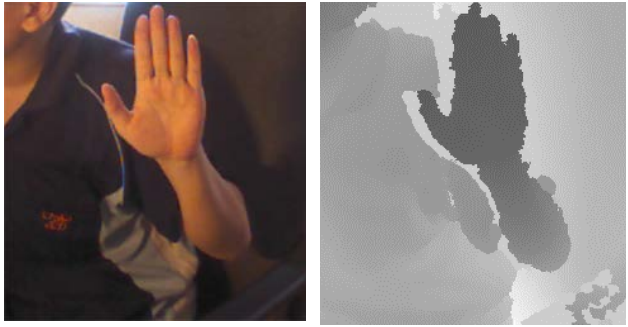

Hình 3. Ảnh quan sát bao gồm: (a) ảnh màu; (b) ảnh độ sâu

## III. GIẢI THUẬT NHẬN DẠNG

Hình 4 trình bày sơ đồ giải thuật nhận dạng được đề xuất trong bài báo. Giải thuật bao gồm 3 giai đoạn chính: trích chọn ảnh quan sát của bàn tay; xây dựng ảnh mô hình giả định của bàn tay tương ứng với góc nhìn quan sát; và tìm tư thế tối ưu bằng giải thuật bầy đàn.

Phần xây dựng ảnh mô hình và ảnh quan sát đã được trình bày ở mục II. Việc tìm tư thế tối ưu được thực hiện qua hai giai đoạn. Giai đoạn thứ nhất là xây dựng hàm mục tiêu để đánh giá sự sai khác giữa ảnh quan sát và ảnh mô hình. Qua đó, chuyển bài toán nhận dạng thành bài toán tối ưu. Giai đoạn thứ hai là giải bài toán tối ưu sử dụng giải thuật bầy đàn. Chi tiết mỗi bước như sau.

### A. Xây dựng hàm mục tiêu

Giả sử có ảnh quan sát $O$, mục tiêu là cần tìm một bộ 26 tham số động học của bàn tay $(q_i^{kq}, q_c^{kq})$ sao cho ảnh mô hình $h_{kq}$ tạo bởi bộ tham số này giống với ảnh quan sát $O$ nhất. Tiêu chí để so sánh sự sai khác giữa ảnh mô hình và ảnh quan sát được xây dựng theo [2] như sau.

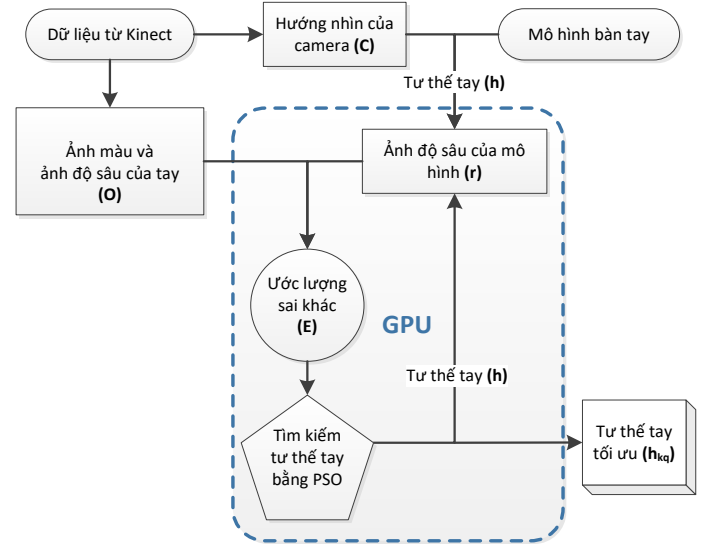

Hình 4. Sơ đồ giải thuật nhận dạng

Xét một ảnh mô hình $h$ bất kì, bằng phép chiếu hình học lên mặt phẳng quan sát với thông tin về tiêu cự và góc nhìn của camera $C$, ta thu được ảnh độ sâu $r_d(h, C)$. Ảnh độ sâu này sau đó được so sánh với ảnh độ sâu quan sát $O_d$ để tìm ảnh tương quan nhị phân $r_m(h, C)$. Quy tắc tính ảnh tương quan như sau:

Giá trị của mỗi điểm ảnh của $r_m(h, C)$ bằng "1" khi tại vị trí đó sai khác giữa $r_d(h, C)$ và $O_d$ nhỏ hơn một khoảng $d_m$ hoặc tại đó $O_d$ không xác định; trong các trường hợp còn lại, giá trị của $r_m(h, C)$ bằng "0".

Ảnh tương quan này sau đó tiếp tục được so sánh với ảnh màu $O_s$ để loại bớt những vùng độ sâu không thích hợp. Kết quả dẫn đến hàm đánh giá sai khác của toàn bộ mô hình như sau:

$$D(O, h, C) = \frac{\sum \min(|o_d - r_d|, d_M)}{\sum(o_s \vee r_m)} + \lambda\left(1 - \frac{2\sum(o_s \wedge r_m)}{\sum(o_s \wedge r_m) + \sum(o_s \vee r_m)}\right) \quad (4)$$

trong đó $\vee$ kí hiệu phép HOẶC lôgic; $\wedge$ kí hiệu phép VÀ lôgic; $d_M$ là hằng số dương giới hạn khác biệt về độ sâu; $\lambda$ là hằng số chuẩn hóa sai khác diện tích; tổng $\Sigma$ được tính trên toàn bộ các điểm ảnh.

Xét về mặt ý nghĩa, tỉ số: $\dfrac{\sum \min(|o_d - r_d|, d_M)}{\sum(o_s \vee r_m)}$ thể hiện sự sai khác về độ sâu giữa ảnh quan sát và ảnh mô hình; còn tỉ số $\dfrac{2\sum(o_s \wedge r_m)}{\sum(o_s \wedge r_m) + \sum(o_s \vee r_m)}$ thể hiện sự sai khác về diện tích giữa hai ảnh. Nói cách khác, một tư thế bàn tay $h$ được xem là nghiệm cần tìm nếu ảnh mô hình tạo bởi nó có sự sai khác về độ sâu và về diện tích với ảnh quan sát là nhỏ nhất.

Để loại trừ những tư thế bàn tay vô lý ví dụ như ngón trỏ và ngón giữa xuyên qua nhau, một lượng $\lambda_k . kc(h)$ được thêm vào để tăng giá trị sai khác trong những trường hợp trên. Kết quả là hàm mục tiêu sau cùng được biểu diễn như sau:

$$E(h,O) = D(O,h,C) + \lambda_k . kc(h) \quad (5)$$

trong đó các tham số cho $D(O,h,C)$ và $E(h,O)$ được chọn như sau: $d_m = 1\,cm$, $d_M = 4\,cm$, $\lambda = 20$, $\lambda_k = 10$ và $kc(h) = \sum_{p \in Q} -\min(\phi(p,h),0)$ với $Q$ là 3 cặp ngón tay không tính ngón cái và $\phi$ biểu diễn sự sai khác về góc giữa 2 ngón tay trong mỗi cặp.

Với hàm mục tiêu (5), bài toán nhận dạng trở thành bài toán tối ưu trong đó cần tìm 26 tham số của tư thế $h$ để $E(h,O)$ cực tiểu. Để giải bài toán này, chúng tôi sử dụng phương pháp tối ưu bầy đàn.

*B. Nhận dạng sử dụng phương pháp tối ưu bầy đàn*

Về lý thuyết, rất khó để tìm lời giải tường minh cho phương trình (5). Thay vào đó, các phương pháp giải thống kê thường được sử dụng như giải thuật Powell [6], giải thuật Nelder – Mead [7], hay giải thuật di truyền [8]. Trong bài báo này, chúng tôi sử dụng phương pháp tối ưu bầy đàn nhờ tốc độ hội tụ nhanh và đơn giản trong cài đặt [9].

Giải thuật bầy đàn giải bài toán tối ưu bằng cách tạo ra một tập hợp gồm $n$ phần tử, mỗi phần tử di chuyển và tiến hóa qua mỗi bước để rồi cuối cùng hội tụ tại điểm tối ưu. Ban đầu, các phần tử được gán một vị trí và vận tốc ngẫu nhiên. Sau đó, tại mỗi bước, mỗi phần tử cập nhật vị trí tốt nhất của nó, $P_k$, và vị trí tốt nhất của cả đàn, $G_k$. Gọi $x_k$ và $v_k$ lần lượt là vị trí và vận tốc hiện tại của mỗi phần tử. Khi đó, vị trí và vận tốc tiếp theo của phần tử đó được cập nhật như sau:

$$v_{k+1} = w(v_k + c_1 r_1 (P_k - x_k) + c_2 r_2 (G_k - x_k)) \quad (6)$$

$$x_{k+1} = x_k + v_{k+1} \quad (7)$$

với $w$ là hệ số giảm vận tốc, $c_1$ là hằng số đặc trưng cho yếu tố cá thể, $c_2$ là hằng số đặc trưng cho yếu tố bầy đàn, $r_1$ và $r_2$ là hai biến ngẫu nhiên phân phối đều trong khoảng [0,1]. Phương trình (6) và (7) hàm ý mỗi phần tử sẽ di chuyển ngẫu nhiên nhưng có khuynh hướng tiến về vị trí tốt nhất của cả đàn và vị trí tốt nhất mà nó đã đi qua. Tương quan giữa yếu tố bầy đàn và yếu tố cá thể được thể hiện qua các hệ số $c_1$ và $c_2$.

Áp dụng vào bài toán nhận dạng, vị trí của mỗi phần tử được định nghĩa là vectơ 26 chiều ứng với 26 tham số động học của bàn tay hay chính là tư thế $h$ của bàn tay. Vận tốc được định nghĩa là vectơ 26 chiều thể hiện sự thay đổi tư thế của bàn tay qua mỗi bước. Khi khởi tạo, vị trí của mỗi phần tử được gieo ngẫu nhiên tạo thành các tư thế $h_1, h_2, \ldots h_n$. Vận tốc ban đầu được đặt bằng 0. Từ phương trình (5), giá trị của hàm mục tiêu $E(h_i, O)$ được tính cho mỗi tư thế. Từ đó, vị trí tốt nhất của mỗi phần tử $P_k$ và vị trí tốt nhất của cả đàn $G_k$ được xác định. Vận tốc của mỗi phần tử ở thế hệ tiếp theo sau đó được xác định bởi phương trình (6) và vị trí tiếp theo được xác định bởi phương trình (7). Trải qua các bước tiến hóa, vị trí ( hay tư thế bàn tay) của mỗi phần tử sẽ tiến dần tới tư thế thực quan sát bởi camera. Thuật toán dừng khi sai số hàm mục tiêu nhỏ hơn giá trị đặt hoặc số bước tiến hóa đạt tới giá trị tối đa cho phép.

Trong hệ thống của chúng tôi, số phần tử của đàn được đặt là 64. Không gian tìm kiếm được giới hạn bởi khoảng giá trị của các phần tử theo bảng 1 và bảng 2. Điều kiện dừng là khi giá trị hàm mục tiêu nhỏ hơn 1.0 hoặc số bước tiến hóa đạt 30. Các hệ số của phương trình (6) được đặt như sau: $c_1 = 2.8$, $c_2 = 1.3$, và $w = 2 / \left|2 - \psi - \sqrt{\psi^2 - 4\psi}\right|$ với $\psi = c_1 + c_2$.

Trong quá di chuyển theo giải thuật PSO, do số chiều lớn nên các vị trí đốt ngón tay thường bị kẹt tại các đỉnh tối ưu cục bộ thay vì tiến tới đỉnh tối ưu toàn cục. Để giải quyết vấn đề này, các phần tử được tạo đột biến (mutation). Cứ sau 3 bước tiến hóa, một nửa số phần tử kém nhất trong đàn được gieo lại ngẫu nhiên 20 chiều tương ứng với các tham số góc của các đốt ngón tay.

## IV. TĂNG TỐC THUẬT TOÁN SỬ DỤNG KHỐI XỬ LÝ ĐỒ HỌA GPU

Do không gian tìm kiếm 26 chiều, giải thuật bầy đàn phải sử dụng tới 64 phần tử tiến hóa qua 30 thế hệ dẫn đến yêu cầu lớn về số lượng phép tính mà nếu xử lý tuần tự bằng CPU sẽ không đảm bảo yếu tố thời gian thực. Để giải quyết vấn đề này, chúng tôi tận dụng khả năng xử lý song song của khối xử lý đồ họa GPU.

Khối xử lý đồ họa GPU, nằm trong cạc đồ họa của máy tính, là một chíp bổ trợ được thiết kế để hỗ trợ CPU trong các tác vụ đồ họa. Do đặc điểm xử lý đồ họa, khối GPU được thiết kế gồm nhiều nhân xử lý (256 trong hệ của chúng tôi) để tính toán song song. Vì vậy, mặc dù mỗi lõi vi xử lý của GPU có năng lực xử lý kém hơn so với CPU, nhưng khi thực hiện song song trên tất cả các lõi thì GPU lại cho kết quả vượt trội.

Để lập trình song song trên GPU, hai nền tảng phổ biến hiện nay là CUDA của hãng Nvidia [10] và OpenCL của tổ chức Kronos [11]. Nền tảng CUDA có ưu điểm dễ cài đặt nhưng chỉ hỗ trợ cạc đồ họa của Nvidia. OpenCL, mặt khác, ra đời sau nhưng là chuẩn mở hỗ trợ tất cả các loại cạc đồ họa nên được chúng tôi sử dụng trong nghiên cứu này. Hình 5 trình bày quá trình tính toán song song cài đặt cho giải thuật bầy đàn. Theo đó, tính toán song song được chúng tôi thực hiện *giữa các phần tử* và *trong mỗi phần tử* của đàn, cụ thể gồm 3 bước như sau:

- *Bước 1:* Mỗi phần tử của đàn được cấp một vùng nhớ riêng trên GPU. Vị trí của mỗi phần tử chính là một tư thế $h$ của bàn tay. Bằng thư viện đồ họa OpenGL [12] và mô hình bàn tay định nghĩa trong phần II.A, một ảnh mô hình 3 chiều của bàn tay được tạo ra với tư thế $h$. Bằng phép chiếu hình học với thông tin hướng nhìn $C$ và thông số camera đã biết, ta tính được ảnh độ sâu $r$ từ ảnh mô hình. Với 64 phần tử của đàn, ta tạo được 64 ảnh độ sâu $r_1, r_2, \ldots, r_{64}$ để dùng cho bước tính hàm mục tiêu tiếp theo. Toàn bộ tiến trình trên và cả những tiến trình ở các bước tiếp theo được thực hiện đồng thời cho 64 phần tử trên 64 vùng nhớ riêng của GPU. Vì vậy, giải thuật bầy đàn được song song hóa giữa các phần tử.

- *Bước 2:* Bây giờ, với mỗi phần tử, ta cần tính giá trị sai khác $D(O,h,C)$ theo phương trình (4) để từ đó tính giá trị hàm mục tiêu $E(h,O)$ theo phương trình (5). Phương trình (4) yêu cầu cần phải thực hiện các phép tính $|o_d - r_d|$, $O_s \wedge r_m$, $O_s \vee r_m$ với từng điểm ảnh. Với độ phân giải 640x480 của Kinect, số điểm ảnh của vùng bàn tay khi đó là rất lớn và không phù hợp cho tính tuần tự. Song song hóa các phép tính này do đó cũng cần được thực hiện. Ý tưởng của chúng tôi là áp dụng các phép toán trực tiếp giữa hai vùng nhớ thay vì lần lượt cho mỗi ô nhớ. Cụ thể, ảnh quan sát màu $O_s$ và ảnh độ sâu $O_d$ được chuyển từ CPU vào bộ nhớ của GPU. Các vùng nhớ này sau đó được sao chép ra 64 vùng tương ứng với số phần tử của đàn. Các phép tính khi đó được thực hiện cho đồng thời tất cả điểm ảnh trong các vùng nhớ lưu $O_s$, $O_d$ và $r_i$.

OpenCV 2.4.9 [15], thư viện đồ họa OpenGL 4.3 [12] và thư viện tính toán song song trên GPU OpenCL 1.2 [11]. Giải thuật bầy đàn được thực hiện với 64 phần tử tiến hóa qua 30 thế hệ trong đó một nửa số phần tử được đột biến cứ mỗi 3 thế hệ.

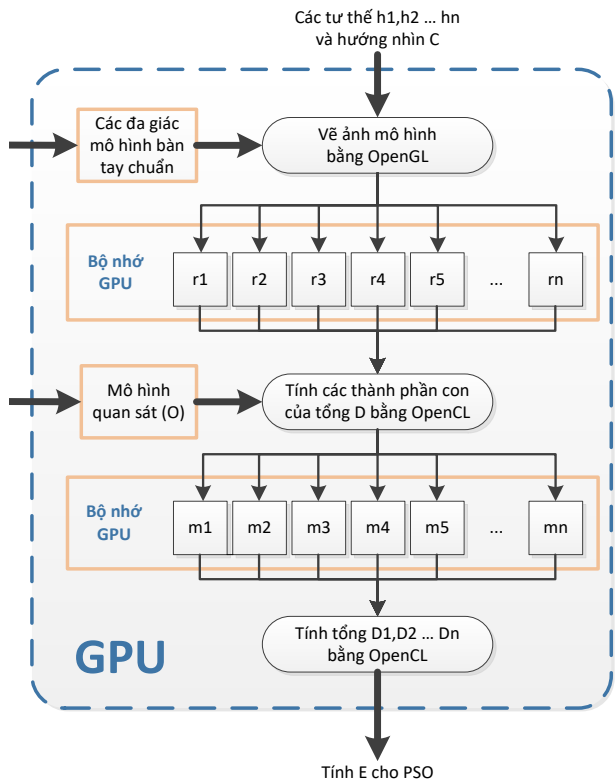

Hình 5: Sơ đồ khối quy trình tính toán trên GPU

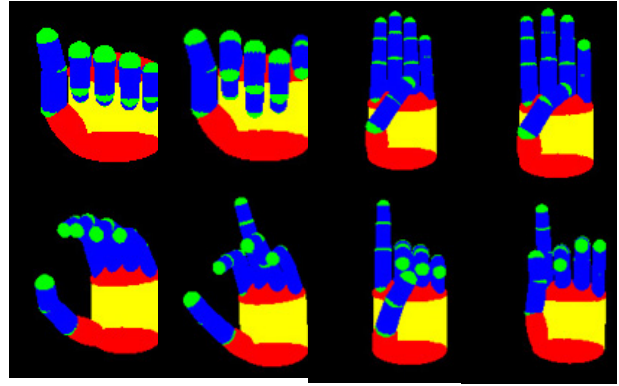

Hình 6: Kết quả nhận dạng 26 bậc tự do bàn tay với 4 tư thế trong đó bên trái là ảnh quan sát và bên phải là ảnh nhận dạng

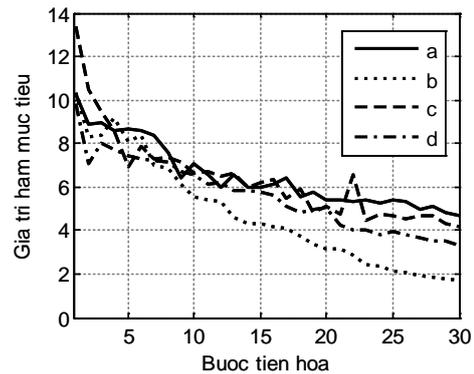

Hình 7: Biến thiên giá trị hàm mục tiêu theo bước tiến hóa với 4 tư thế ứng với các chữ cái "a", "b", "c", "d"

- *Bước 3:* Kết quả của mỗi phép tính ở bước 2 được lưu ở một vùng nhớ có kích thước bằng với vùng nhớ $r_i$. Để tính $D(O,h,C)$, ta còn cần phải tính tổng giữa các phần tử của từng vùng nhớ này (phương trình (4) ). Để tận dụng triệt để khả năng song song của GPU, chúng tôi tiếp tục sử dụng giải thuật tính tổng theo kĩ thuật kim tự tháp [13] để song song hóa quá trình tính tổng. Kết quả là, với cấu hình GPU gồm 256 nhân, thanh ghi kích thước 128 bit, kiểu số thực 32 bit, có 1024 phép tính tổng sẽ được thực hiện đồng thời. Giá trị cuối cùng của $D(O,h,C)$ sau đó được chuyển sang bộ nhớ của CPU để tiếp tục các bước của giải thuật bầy đàn.

Như vậy, bằng cách sử dụng khối xử lý đồ họa GPU, giải thuật bầy đàn PSO đã được song song hóa hoàn toàn. Qua thử nghiệm của chúng tôi, phương pháp này đã giảm thời gian xử lý một khung hình 450 lần từ 6 phút xuống còn 0.8 giây và nhờ đó đảm bảo yếu tố thời gian thực.

## V. MÔ PHỎNG VÀ THỰC NGHIỆM

Để đánh giá hiệu quả của phương pháp đề xuất, chúng tôi đã tiến hành mô phỏng với dữ liệu tổng hợp và thực nghiệm với dữ liệu thật. Hệ thống của chúng tôi được cài đặt trên máy tính xách tay có cấu hình CPU Intel i7 740qm, 8 Gb RAM, cạc đồ họa GPU AMD HD5870m có năng lực xử lý 1,12 TFlops cùng 1 Gb bộ nhớ. Chương trình phần mềm được viết trên nền tảng Visual C++ 2010 [14] kết hợp với thư viện xử lý ảnh

### A. Mô phỏng

Mô phỏng được thực hiện với mục đích đánh giá và hoàn thiện giải thuật trước khi áp dụng với dữ liệu thực. Chi tiết như sau.

*1) Cài đặt mô phỏng*: Để thực hiện mô phỏng, một tư thế tay tùy chọn được đưa vào hệ thống để tạo ảnh mô hình $h_{ref}$. Giả thiết có hướng nhìn $C$, thông tin camera và mặt phẳng quan sát, khi đó ảnh màu và ảnh độ sâu được tạo ra từ ảnh mô hình bằng phép chiếu hình học. Các ảnh này được giả thiết như là ảnh quan sát màu $O_s$ và ảnh độ sâu $O_d$ thu được từ camera. Với các ảnh giả thiết này, giải thuật nhận dạng có thể thực hiện như với dữ liệu thực.

*2) Kết quả mô phỏng*: Hình 6 trình kết quả nhận dạng 26 bậc tự do đối với các tư thế tay tương ứng với 4 chữ cái đầu của bảng chữ cái ngôn ngữ kí hiệu trong đó ảnh bên trái biểu diễn tư thế tay quan sát $h_{ref}$ và ảnh bên phải biểu diễn tư thế tay nhận dạng $h_{kq}$. Có thể nhận thấy giải thuật đã xác định được chính xác 26 bậc tự do của bàn tay với một số tư thế. Trong một số tư thế khác, các bậc tự do gắn với các đốt ngón tay cho kết quả chưa chính xác do các phần tử của đàn bị tắc ở điểm tối ưu cục bộ. Hình 7 trình bày sự thay đổi giá trị hàm mục tiêu qua từng bước tiến hóa. Có thể nhận thấy giá trị hàm mục tiêu giảm khi số bước tiến hóa tăng hay nói cách khác vị trí các phần tử của đàn tiến dần tới tư thế cần tìm. Trung bình, việc nhận dạng một tư thế được thực hiện trong 0.8 giây, trong đó 0,45 giây tiêu tốn cho việc xây dựng ảnh mô hình, ảnh quan sát, ảnh màu và 0,35 giây cho việc tính toán hàm mục tiêu $E$.

*B. Thực nghiệm*

Trong thực nghiệm, ảnh quan sát là dữ liệu thu được từ cảm biến ảnh Kinect phiên bản 1.5 trong điều kiện trong nhà và ánh sáng ổn định. Vị trí bàn tay đặt cách cảm biến trong khoảng từ 0.5 $m$ đến 1.5 $m$.

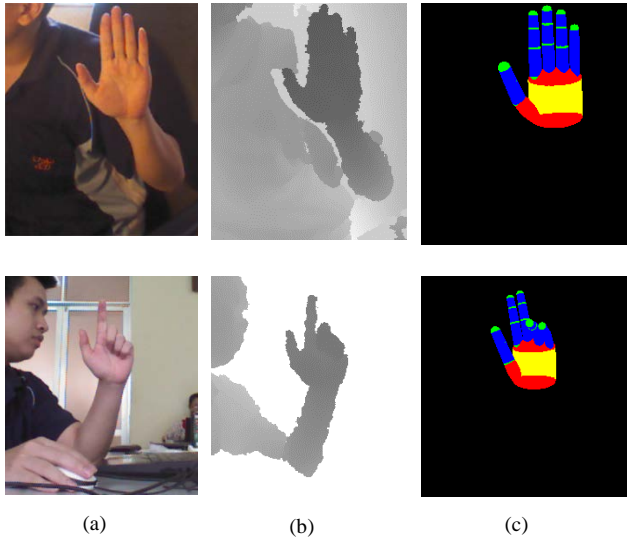

(a)      (b)      (c)

Hình 8: Kết quả thực nghiệm nhận dạng tư thế tay: (a) Ảnh quan sát màu; (b) Ảnh quan sát độ sâu; (c) Ảnh kết quả nhận dạng

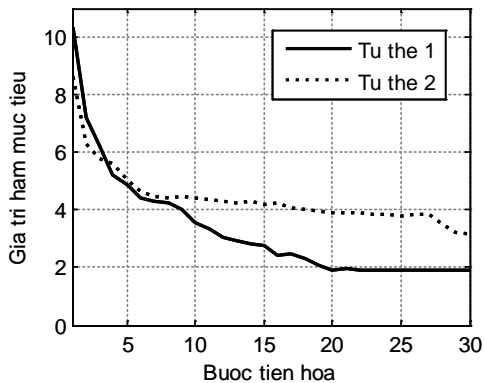

Hình 9: Biến thiên giá trị hàm mục tiêu theo bước tiến hóa với 2 tư thế trong hình 8

Hình 8 trình bày kết quả nhận dạng. Trong cả hai tư thế, 6 bậc tự do của cổ tay và 4 bậc tự do của ngón cái được nhận dạng chính xác. Với bậc tự do của các ngón tay còn lại, chỉ các tư thế duỗi ngón được nhận dạng chính xác. Các tư thế gập ngón do một phần ngón tay bị che khuất nên thường dẫn tới kết quả không chính xác. Đồ thị giá trị hàm mục tiêu (hình 9) cho thấy, việc thay đổi tư thế các đốt ngón tay không dẫn tới sự thay đổi lớn của giá trị hàm mục tiêu và do đó không giúp các phần tử thoát khỏi đỉnh tối ưu cục bộ. Để khắc phục vấn đề này, cần thiết phải cải tiến hàm mục tiêu để nâng cao khả năng phân biệt. Việc này sẽ được thực hiện trong nghiên cứu tiếp theo của chúng tôi.

## VI. KẾT LUẬN

Đóng góp chính của báo cáo là đã xây dựng thành công giải thuật nhận dạng 26 bậc tự do của bàn tay từ khâu thu thập dữ liệu, xây dựng mô hình, xây dựng hàm mục tiêu tới triển khai giải thuật tối ưu bầy đàn. Đặc biệt, việc tính toán song song giải thuật bầy đàn trên khối xử lý đồ họa GPU đã giúp cải thiện đáng kể hiệu năng xử lý của hệ thống mà không yêu cầu phần cứng bổ sung. Nhiều phép mô phỏng và thực nghiệm đã được tiến hành để khẳng định tính đúng đắn của phương pháp. Trong thời gian tiếp theo, hệ thống sẽ tiếp tục được cải tiến theo hai hướng bao gồm: trích chọn thêm đặc trưng của bàn tay để cải thiện khả năng đánh giá của hàm mục tiêu; và tăng cường phần cứng và tối ưu giải thuật song song để có thể xử lý được 30 hình/giây. Trên cơ sở đó, chúng tôi kì vọng xây dựng được các ứng dụng tương tác người – máy tiên tiến như thực tại ảo và thực tại tăng cường.

## TÀI LIỆU THAM KHẢO

### Reference

[1] A. Erol, G. Bebis, M. Nicolescu, R.D. Boyle, X. Twombly, "Vision-based Hand Pose Estimation: A review", J. Computer Vision and Image Understanding, vol.108(1-2), pp.52–73, 2007.

[2] I. Oikonomidis, N. Kyriazis, A. Argyros, "Efficient model-based 3D tracking of hand articulations using Kinect", Proceedings of the British Machine Vision Conference, pp 101.1-101.11, 2011.

[3] H. Ouhaddi, P. Horain, "3D Hand gesture tracking by model registration", International Workshop on Synthetic—Natural Hybrid Coding and Three Dimensional Imaging, 1999.

[4] B. Stenger, P.R.S. Mendonca, R. Cipolla, Model-based 3D tracking of an articulated hand, IEEE Computer Society Conference on Computer Vision and Pattern Recognition 02 (2001) 310.

[5] B. Stenger, Model-based hand tracking using a hierarchical bayesian filter, Ph.D. thesis, Department of Engineering, University of Cambridge, 2004.

[6] W. H. Press, B. P. Flannery, S. A. Teukolsky, W. T. Vetterling, Numerical Recipes in C, Cambridge University Press, 1992.

[7] J. A. Nelder and R. Mead, "A Simplex Method for Function Minimization", Computer Journal, vol. 7, 1965, pp. 308-313.

[8] David Goldberg, *Genetic Algorithms in Search, Optimization and Machine Learning*, Addison-Wesley Professional, 1989.

[9] J. Kennedy, R.C. Eberhart, *Swarm Intelligence*, Morgan Kaufmann, 2001.

[10] Nicholas Wilt, The CUDA Handbook: A Comprehensive Guide to GPU Programming, Addison-Wesley Professional, 1 edition, 2013.

[11] Matthew Scarpino, OpenCL in Action: How to Accelerate Graphics and Computations, Manning Publications, 1 edition, 2011.

[12] Dave Shreiner, Graham Sellers, John M. Kessenich, Bill M. Licea-Kane, OpenGL Programming Guide: The Official Guide to Learning OpenGL, Addison-Wesley Professional, 8 edition, 2013.

[13] L. Williams, Pyramidal parametrics, In ACM SIGGRAPH Computer Graphics, vol. 17, pp 1–11, 1983.

[14] Ivor Horton, Ivor Horton's Beginning Visual C++ 2010, Wrox; 1 edition, 2010.

[15] Samarth Brahmbhatt, Practical OpenCV (Technology in Action), Apress, 1 edition, 2013

[16] Jana Abhijit, Kinect for Windows SDK Programming Guide (Community Experience Distilled), Packt Publishing, 2012.

[17] Manh Duong Phung, Quang Vinh Tran, K. Hara, H. Inagaki and M. Abe, "Easy-Setup Eye Movement Recording System for Human-Computer Interaction," The 2008 IEEE International Conference on Research, Innovation and Vision for the Future (RIVF), pp.292-297, Ho Chi Minh City, Vietnam, 2008